\documentclass[pdflatex, iicol]{sn-jnl}%

\usepackage{graphicx}%
\usepackage{multirow}%
\usepackage{amsmath,amssymb,amsfonts}%
\usepackage{amsthm}%
\usepackage{mathrsfs}%
\usepackage[title]{appendix}%
\usepackage{xcolor}%
\usepackage{textcomp}%
\usepackage{manyfoot}%
\usepackage{booktabs}%
\usepackage{algorithm}%
\usepackage{algorithmicx}%
\usepackage{algpseudocode}%
\usepackage{listings}%
\usepackage[scaled]{helvet} 

\usepackage[font=sf]{caption}
\usepackage{titlesec}
\titleformat*{\section}{\Large\bfseries\sffamily}
\titleformat*{\subsection}{\large\bfseries\sffamily}
\titleformat*{\subsubsection}{\normalsize\bfseries\sffamily}
\usepackage{lipsum}
\usepackage{dblfloatfix}

\theoremstyle{thmstyleone}%
\theoremstyle{thmstyletwo}%

\theoremstyle{thmstylethree}%

\begin{document}

\title[Article Title]{Explainable Part-Based Vehicle Classifier with Spatial Awareness}

\author[1]{\fnm{Andreas} \sur{Caduff}}\email{andreas.caduff@hslu.ch}
\author[1]{\fnm{Klaus} \sur{Zahn}}\email{klaus.zahn@hslu.ch}
\author[1]{\fnm{Jonas} \sur{Hofstetter}}\email{jonas.hofstetter@hslu.ch}
\author[1]{\fnm{Martin} \sur{Rechsteiner}}\email{martin.rechsteiner@hslu.ch}
\author[2]{\fnm{Patrick} \sur{Flaig}}\email{patrick.flaig@sick.ch}

\affil[1]{\orgdiv{Competence Center for Intelligent Sensors and Networks}, \orgname{Lucerne University of Applied Science and Art}, \orgaddress{\street{Technikumstr. 21}, \city{Horw}, \postcode{6048}, \country{Switzerland}}}
\affil[2]{\orgname{SICK AG}, \orgaddress{\street{Erwin-Sick-Str. 1}, \city{Waldkirch}, \postcode{79183 },\country{Germany}}}

\abstract{
In the area of Intelligent Transportation Systems (ITS), fine-grained vehicle classification systems play an essential role. Recently, the authors have presented a novel vision-based classification approach in which standard end-to-end Convolutional Neural Networks (CNNs) have been decomposed into 1) a CNN-based detector for semantically strong vehicle parts, followed by 2) feature construction and 3) final classification by a decision tree. In contrast to conventional CNNs, this allows both easy extensibility to new vehicle categories – without the need to fully retrain the part detector – and an important step towards the interpretability of the model, removing partially the black-box nature inherent to CNNs.

Here we present an important extension of this approach that now incorporates spatial awareness of the vehicle parts: while the feature construction 2) of the previous approach used a binary decision for each feature (present vs. absent), now a full spatial probability map is constructed to condition the presence of each individual part with respect to a given vehicle category. The classification is performed using a softmax regression approach for the overall vehicle probabilities. This method shows a considerably improved robustness against false (part-)detections, a point that is crucial for practical application.

Comparative analyses with a state-of-the-art end-to-end CNN indicate that our part-based methods achieve comparable accuracy, effectively challenging the presumed trade-off between accuracy and explainability. This research represents a significant advance in vehicle classification for ITS and forms the basis for systems that combine high accuracy with intuitive interpretability. 
}


\maketitle

\section{Introduction}\label{introduction}

Traffic monitoring systems are a cornerstone of Intelligent Transportation Systems (ITS), providing important real-time data on vehicle frequencies and categories. This data is an important basis for private and public transport operators to manage toll operations, optimize traffic flow, and enhance traffic safety. Over the last decade, significant progress has been made in camera hardware and image processing algorithms, particularly in machine learning. Vision-based systems are increasingly used in ITS applications, thanks in part to the groundbreaking work of Krizhevsky et al. \cite{krizhevsky2012imagenet} with deep Convolutional Neural Networks (CNN). Now CNNs have established themselves as the state-of-the-art method in vision-based classification, a trend that also applies to ITS applications \cite{liu2022convnet,won2020intelligent}.

With the increasing performance of CNN-based classification systems, the requirements of ITS operators for a more detailed granularity of vehicle categorization are also increasing. Previous research \cite{caduff2021exploring} has shown that vision-based classification using state-of-the-art CNNs can distinguish up to 19 fine-grained categories, meeting the stringent traffic standard TLS8+1 A1 \cite{TLS8+1}. Despite this progress, practical application of CNN-based classification in ITS faces two major challenges. The first challenge (Problem I) lies in the inherent black-box nature of CNNs, which makes it difficult to understand and suppress misclassifications. The second challenge (Problem II) is the need to completely retrain the CNN when modifying existing categories or adding new ones.

To address these challenges, intense research efforts are currently underway. For Problem II, Class- Incremental Learning (CIL) schemes have been proposed, which focus on efficient learning of new classes while avoiding catastrophic forgetting. Within this domain fixed representation approaches are noteworthy, where – generally speaking – the basic feature extraction part (e.g. convolutional layers for CNN) are not updated for each incremental state \cite{belouadah2021comprehensive}. This prioritizes stability over flexibility, which meets the requirements of ITS applications. Problem I is being addressed in the area of Explainable Artificial Intelligence (XAI). Researchers are exploring alternatives to end-to-end CNNs that allow for interpretability and explainability in applications, where a black-box approach is a no-go \cite{zhang2021survey}. Notable strategies in this area include decomposing the process to generate a semantically meaningful representation of the object and employing inherently interpretable classifiers like decision trees \cite{zhang2018visual}.

Building upon these insights, the authors recently proposed a novel approach to achieve both explainability and simple extensibility \cite{caduff2022disentangling}. This was obtained by segmenting the classification process into a CNN-based detection of vehicle parts, a feature construction phase, and a decision tree for the final classification \cite{zhao2017survey}. This part-based classifier achieves the accuracy of traditional end-to-end CNN methods by providing better explainability and easy extensibility. Nevertheless, it still has a significant drawback, which is a pronounced sensitivity to false detections in the vehicle part detector. Because any wrong part detection – be it false positive or false negative – almost inevitably leads to a wrong class determination due to the hard decisions in the final tree classifier. We simulated the case of false part detections by decreasing the detection threshold from the default value 0.5 down to 0.001. This reduces the overall classification accuracy from 98.5\% to less than 93\%, far below the requirements of ITS. Obviously, this represents a problem for real-world ITS applications where robustness is a key issue.

Here we present a new, improved approach that solves these problems through the integration of spatial relationship of vehicle parts. Therefore, so-called part location scores are computed that represent a spatial probability map for the presence of each part for a given vehicle category   and which are integrated into the classification process through a softmax regression. These ideas are based on work by Felzenszwalb et al. \cite{felzenszwalb2009object} – and subsequent studies on fine-grained classification such as \cite{zhang2014part} – which show that spatial relationship of vehicle parts may represent a key element in object recognition. We will demonstrate, that this improvement makes the overall classifier almost insensitive to the threshold of the part detector and thus improves its robustness to false detection considerably. Nevertheless, the classification performance remains comparable – or partially even improves – with respect to both the CNN and our previous part-based method \cite{caduff2022disentangling}.

The contributions of this paper are as follows:

\begin{itemize}
    \item We extend our previous approach with a spatial awareness of the vehicle parts and obtain a considerable improvement in terms of robustness for real-word applications.
    \item Nevertheless, our improved model provides simple extensibility to new classes and – partial – explainability but maintains the detection accuracy of a state-of-the-art CNN.
    \item To our knowledge this represents in terms of size and granularity the largest study performed in the context of fine-grained vehicle detection including a thorough validation in a practical application.
\end{itemize}

Finally, in the present journal paper we provide considerably more details e.g. on the preparation of the training data or the overall system architecture than presented in our previous conference contributions.

\section{Related Work}\label{related}

\begin{figure*}[]
    \centering
    \includegraphics[width=130mm]{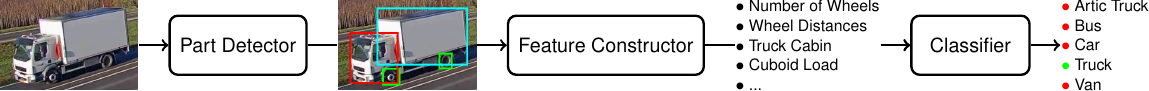}
    \caption{General workflow of the part-based classification approach: 1) using a CNN, semantically meaningful vehicle parts are detected, which are the basis for 2) the feature construction, followed by 3) a conceptually simple and explainable classifier.}\label{fig_concept}
\end{figure*}

The landscape of vehicle classification within ITS has seen a remarkable evolution, driven by advancements in several interconnected research domains. To properly situate our work within this broader scientific context, it is important to look at the most important developments in three key areas: fine-grained classification, XAI, and CIL. Each of these fields uniquely contributes to vehicle classification, providing valuable insights and methodologies that directly address the challenges identified in our research.

\subsection{Fine-grained Classification}
Fine-grained image classification, supported by deep learning, has gained substantial attention in recent years. This area focuses on distinguishing objects from closely related subcategories, such as different bird species or car models. These tasks are particularly challenging due to minimal visual inter-class differences and significant intra-class appearance variations caused by scale, viewpoint changes, complex backgrounds, and occlusions \cite{zhao2017survey}. There are three primary approaches to fine-grained classification:

\textbf{Holistic Approach:} This method employs state-of-the-art ent-to-end CNNs to learn discriminative representations of objects for classification. Two popular strategies are using a single CNN \cite{huttunen2016car,adu2017automated,zhuo2017vehicle,hu2017location,butt2021convolutional,gholamalinejad2021vehicle,sasongko2019indonesia,armin2020vehicle} or an ensemble of CNNs \cite{jung2017resnet,liu2017ensemble,kim2017vehicle,jagannathan2021moving,taek2017deep,hedeya2020super,theagarajan2017eden,rachmadi2018single}.

\textbf{Attention-Based Approach:} Inspired by the human visual system, this approach emphasizes salient features, especially useful in the context of cluttered images \cite{hu2017deep,chen2016visual,yu2020cam,boukerche2021novel}.

\textbf{Part-Based Approach:} Here, semantic part localization and extraction of their discriminative features play a crucial role. This strategy involves either manually defining object parts \cite{he2015recognition,wang2020multi,huang2016fine,liao2022multi} or learning them during training \cite{fang2016fine,krause2014learning,tian2018selective,lu2022novel}, with some methods leveraging the spatial relationships between these parts \cite{biglari2017cascaded,liao2015exploiting,hu2015learning,xiang2019global1,xiang2019global2}. Another variant involves fitting a 3D model to the object to locate discriminative parts \cite{krause20133d,lin2014jointly,sochor2018boxcars,rui2020geometry}.

\subsection{Explainable Artificial Intelligence (XAI)}
Despite the success of Deep Neural Networks (DNNs) there are considerable concerns about their opacity. This lack of interpretability is a major problem in high-risk scenarios where people may be harmed, financial losses may occur or legal consequences may arise \cite{zhang2021survey}. Approaches like those by Chen et al. \cite{chen2019looks}, Rymarczky et al. \cite{rymarczyk2022interpretable}, Nauta et al. \cite{nauta2021neural}, and Donnelly et al. \cite{donnelly2022deformable} propose various network architectures that dissect images into prototypical parts for classification, employing simple linear models or decision trees for enhanced interpretability.

\subsection{Class-Incremental Learning (CLI)}
CIL remains a critical challenge in AI, particularly the issue of catastrophic forgetting \cite{belouadah2021comprehensive}. Various methods have been developed to address this problem, such as Yan et al. \cite{yan2021dynamically}, Hayes et al. \cite{hayes2020lifelong,hayes2020remind}, Zhang et al. \cite{zhang2021few}, Belouadah et al. \cite{belouadah2018deesil}, Petit et al. \cite{petit2023fetril}, and Li et al. \cite{li2022few}. These approaches focus on techniques like freezing learned representations, employing decoupled learning strategies, and using graph models or linear discriminant analysis to incorporate new class knowledge without forgetting old information.

\subsection{Approaches related to our Work}
There are methodologies in the literature that show similarities with our previous and current work, and these should be briefly highlighted. He et al. \cite{he2019deep} proposed a truck classification approach where a CNN-based vehicle detection is combined with a – handcrafted – feature extractor and a subsequent decision tree. However, the approach is limited to trucks and does not use any vehicle parts. Ma et al. \cite{ma2023image} combined self-supervised learning with wheel positional features but relied mainly on abstract, non-explainable features. Wang et al. \cite{wang2015probabilistic} and Li et al. \cite{li2020multi} applied Gaussian models to account for the spatial relationship of vehicle parts for car localization, a related but different task from fine-grained vehicle classification.

While these are partial solutions to the three problems mentioned, our approach uniquely addresses them as a whole – to the best of our knowledge a first in the field of ITS vehicle classification. This holistic treatment of fine-grained classification, XAI, and CIL within a single frame-work sets our work apart and makes it a significant contribution to the field.

\section{Methodology}\label{method}

As shown in Figure \ref{fig_concept}, our general workflow comprises three main steps: 1) the detection of semantically strong vehicle parts, 2) the construction of features from these parts, and 3) the use of a simple, interpretable classifier. While step 1) is identical for our previous \cite{caduff2022disentangling} and the present approach, steps 2) and 3) will differ as explained in detail below. Nevertheless, this overall architecture is the key to overcoming the problems of conventional CNN-based classification methods.

We focus on two primary problems: first, explainability, which is achieved through the detection of meaningful vehicle parts and the use of an understandable classifier, and second, simple extensibility, which is achieved by keeping the part detector fixed and only adapting the feature construction and classifier.

In the following sections we first describe the baseline CNN architecture, which provides the benchmark for performance comparison. Then, we briefly recall our previous decision tree approach \cite{caduff2022disentangling} as basis for the presentation of our current model. A significant focus will be on the improvement obtained through the integration of the spatial relationship between the parts in combination with the softmax regression approach.
 
In addition, we outline the preparation of the datasets, consisting of the vehicle categories and their parts, as well as the efficient annotation process to provide a sufficiently large dataset, which is essential for a successful training and good generalization capabilities of the model. We finish with a thorough evaluation of our approach including comparison to our prior work and the baseline CNN.

\subsection{Dataset}\label{data}

Our research on image-based vehicle classification for traffic monitoring systems is based on an extensive dataset tailored to our specific needs. The creation of this dataset was necessary because there were no public datasets available that provided the required level of 19 fine-grained categories (c.f. Table \ref{tab_categories}), annotated vehicle parts, and specific viewpoints \cite{caduff2021exploring,caduff2022disentangling}.

\textbf{Data Collection Process and Vehicle Category Determination:} The images were acquired from – wide-angle – roadside cameras and cropped to the relevant regions of interest determined by a state-of-the-art object detector of type YOLO \cite{redmon2016you}. The determination of the vehicle category was significantly simplified through the use of a roadside installed laser-system providing a first hypothesis for the fine-grained vehicle type. In subsequent step, these class annotations where manually verified and – if necessary – corrected to ensure a reliably correct annotation even for complex categories \cite{caduff2021exploring}.

\textbf{Vehicle Part Definition and Annotation:} 
This obviously represented the most challenging and time-consuming part of the data preparation. Guideline to the part definition was the goal to identify conceptually simple yet semantically strong parts that distinctly represent all vehicle categories. The final selection is represented in Table \ref{tab_parts}. Prior to the training of the part detector its “completeness” , i.e. the possibility to distinguish all required 19 fine-grained categories, was tested theoretically using a decision tree. 

\begin{figure}[]
    \centering
    \includegraphics[width=74mm]{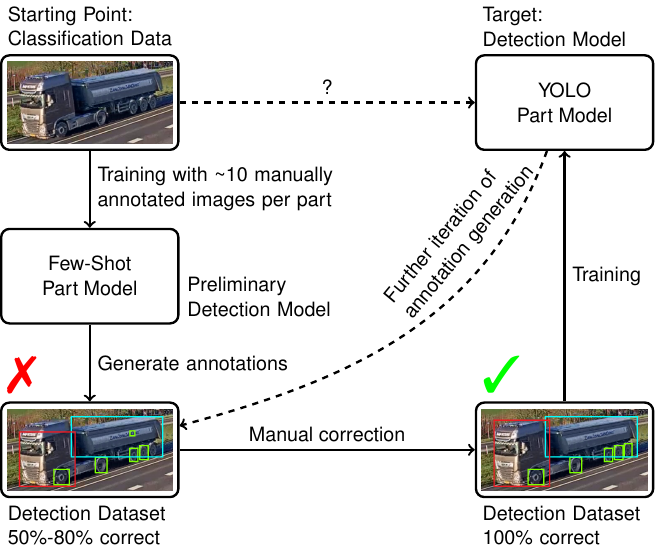}
    \caption{Workflow for the efficient annotation of vehicle parts in our image dataset. A few-shot based approach is used to train an initial model. Then, iteratively erroneous annotations are manually corrected (deleting the false positives and correcting the false negatives) to increase the size of the training dataset to improve the model, etc.}\label{fig_data_acq}
\end{figure}

For the actual vehicle part annotation a semi-supervised approach as illustrated in Figure \ref{fig_data_acq} was applied. As bootstrap an initial part detector based on a few-shot method \cite{wang2020generalizing} was applied. Therefore a FSS1000 model \cite{li2020fss} was fine-tuned with a handful (~10) of hand-annotated images for each of the 16 vehicle parts. 

This approach already obtained surprisingly good detection results, which were manually corrected in a subsequent step. Therefore, images containing the bootstrap model’s annotations were divided into three categories: images with fully correct annotation (”correct”), images with fully correct annotation but additional false positives (”correct with false positives”) and images with missing or incorrect annotations (”incorrect”). Removing false positives from the ”correct with false positives” category was very efficient so that the manual correction was mainly limited to the ”incorrect” category. With this procedure, the manual annotation effort was reduced by 50\% to 80\%, depending on the vehicle part, and yielded a fully corrected annotated dataset, which served as training data for the final part detection model (YOLO in Figure \ref{fig_data_acq}).  

\textbf{Dataset Enhancement and Refinement:} The final datasets shown in Table \ref{tab_categories} and Table \ref{tab_parts} include 19 fine-grained vehicle categories and annotations for 16 distinct vehicle parts, with the respective sample size ranging from dozens to hundreds of samples. This actually represents a minor extension to the set presented in our previous work \cite{caduff2022disentangling}: now, very rare categories (Tractor Truck, Truck Car Transporter Empty/Loaded) and the corresponding vehicle parts (Roof/Support Truck Car Transporter) are included.

\begin{table}[]
\caption{Vehicle categories and corresponding number of samples}\label{tab_categories}
\begin{tabular}{lrr}
\toprule
                Vehicle Category &     Samples &                      Sample Image \\
\midrule
                    Artic Truck &           458 &   \includegraphics[height=5mm]{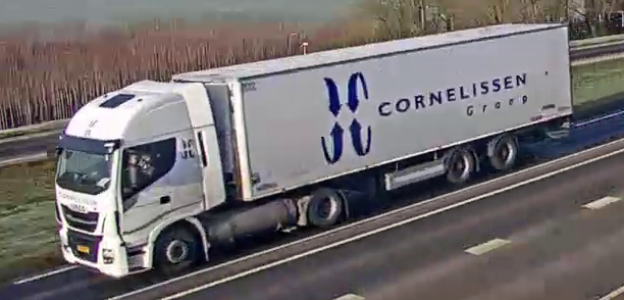} \\
            Artic Truck Dumptor &            86 &   \includegraphics[height=5mm]{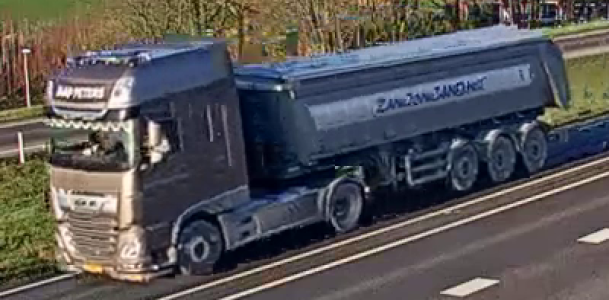} \\
        Artic Truck Low Loaded &            48 &    \includegraphics[height=5mm]{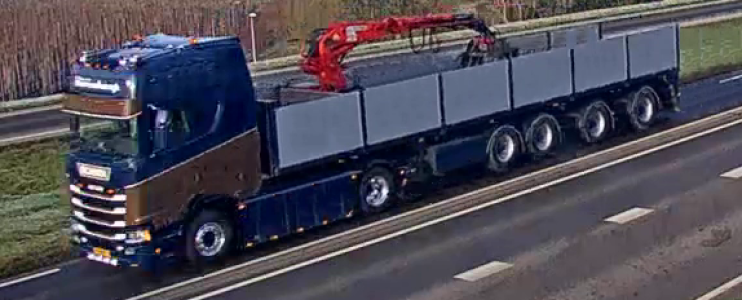} \\
            Artic Truck Tanker &            50 &    \includegraphics[height=5mm]{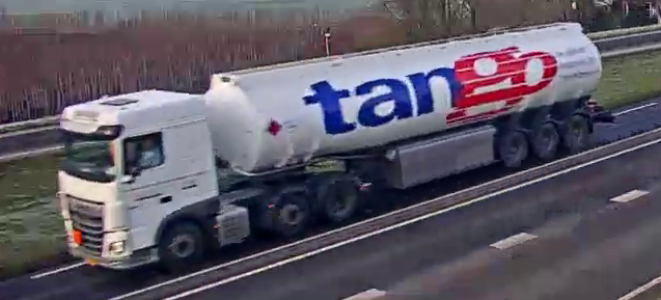} \\
                    Artic Van &            46 &     \includegraphics[height=5mm]{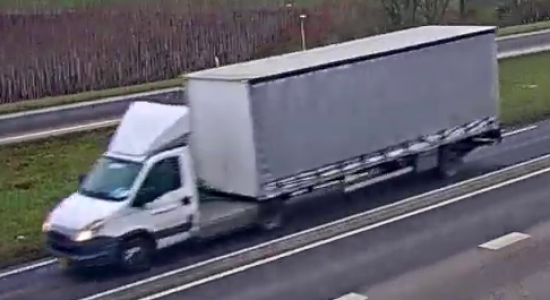} \\
                            Bike &           272 &  \includegraphics[height=5mm]{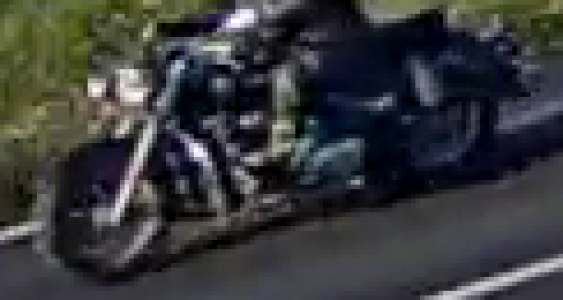} \\
                            Bus &           213 &   \includegraphics[height=5mm]{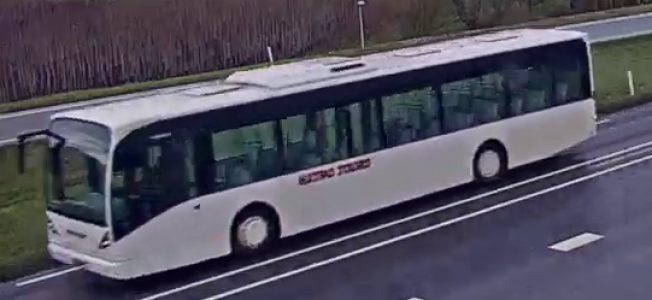} \\
                    Camper Van &           255 &    \includegraphics[height=5mm]{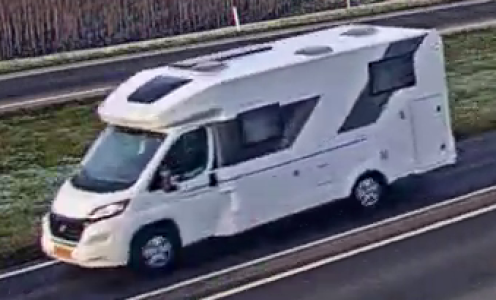} \\
                            Car &          1258 &   \includegraphics[height=5mm]{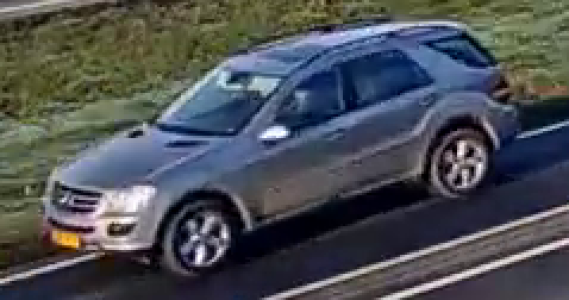} \\
                Tractor Truck &            23 &     \includegraphics[height=5mm]{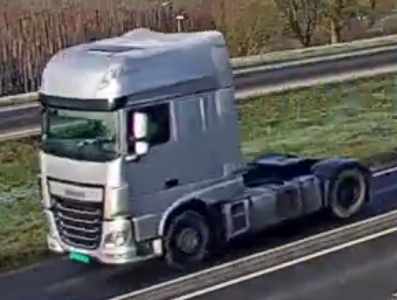} \\
                        Trailer &           495 &   \includegraphics[height=5mm]{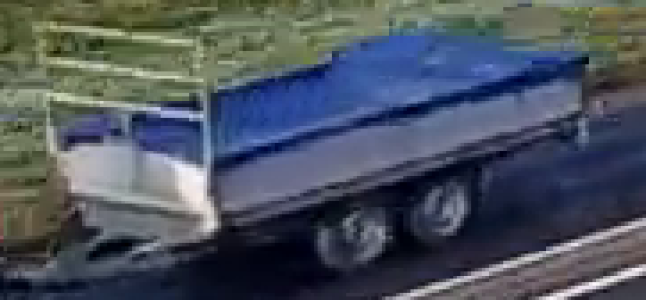} \\
                        Truck &           313 &     \includegraphics[height=5mm]{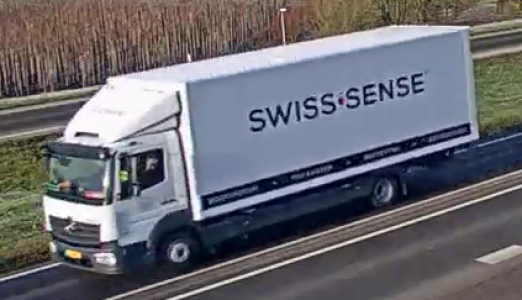} \\
    Truck Car Transporter Empty &            46 &   \includegraphics[height=5mm]{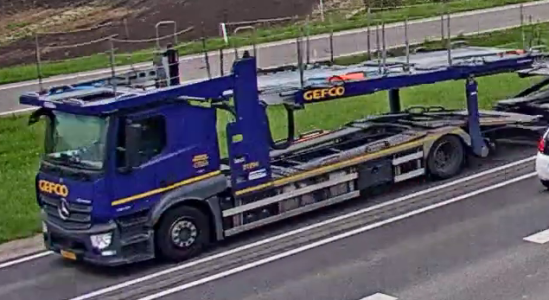} \\
    Truck Car Transporter Loaded &            40 &  \includegraphics[height=5mm]{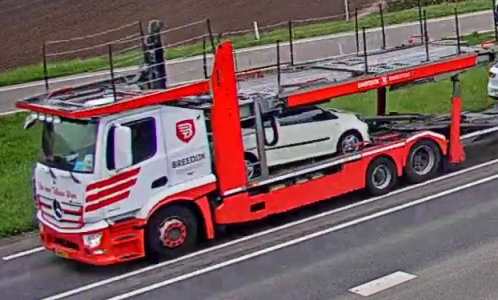} \\
                Truck Dumptor &           238 &     \includegraphics[height=5mm]{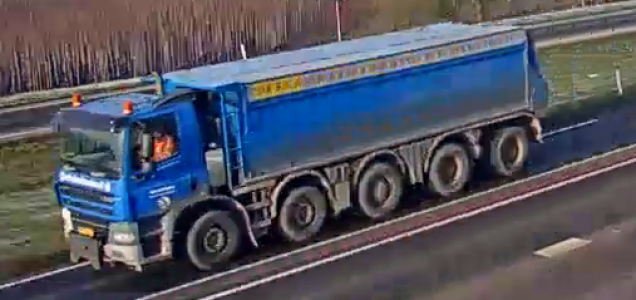} \\
                Truck Low Loaded &           139 &  \includegraphics[height=5mm]{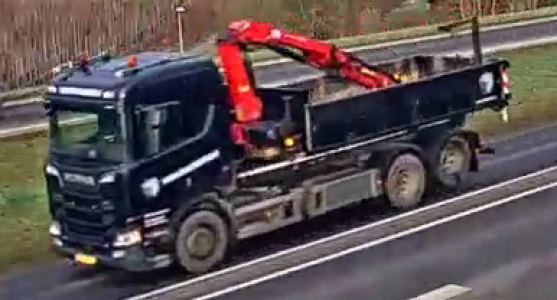} \\
                    Truck Tanker &           190 &  \includegraphics[height=5mm]{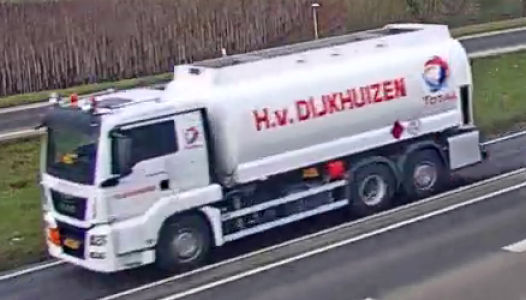} \\
                            Van &           336 &   \includegraphics[height=5mm]{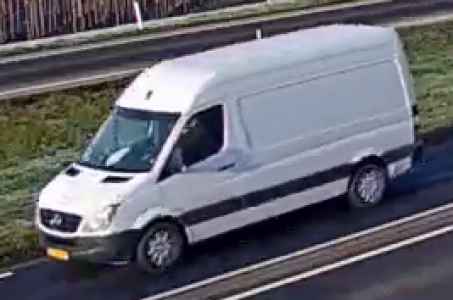} \\
                    Van Pickup &           273 &    \includegraphics[height=5mm]{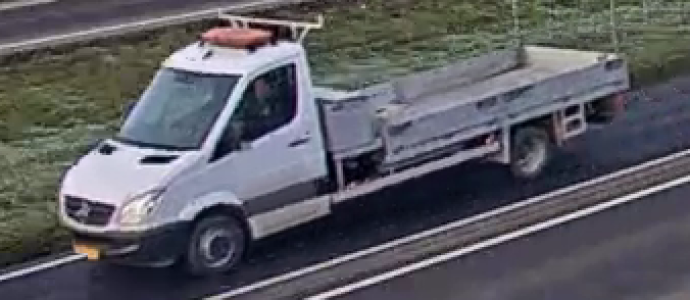} \\
\midrule
                \textbf{Total} &  \textbf{4779} &                                  \\
\bottomrule
\end{tabular}
\end{table}

\begin{table}[]
\caption{List of vehicle parts and corresponding number of annotated parts (green masks)}\label{tab_parts}
\begin{tabular}{lrr}
\toprule
                    Vehicle Part &  Objects &                      Sample Image \\
\midrule
                        Body Bike &            274 &  \includegraphics[height=5mm]{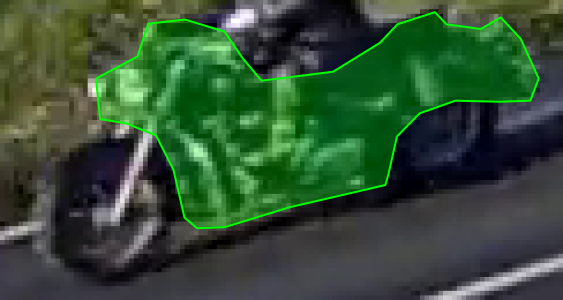} \\
                        Front Bus &            213 &  \includegraphics[height=5mm]{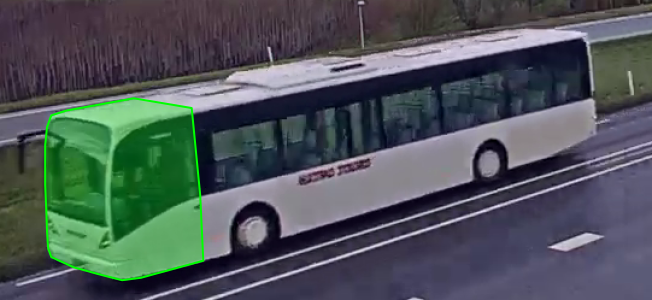} \\
                        Front Car &           1302 &  \includegraphics[height=5mm]{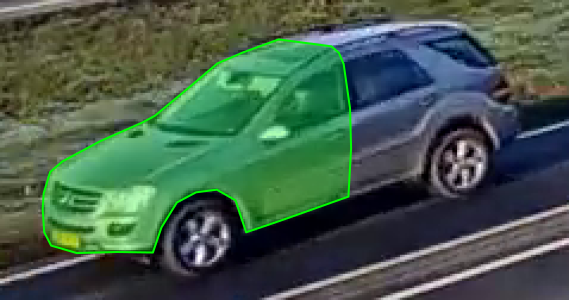} \\
                    Front Truck &           1631 &    \includegraphics[height=5mm]{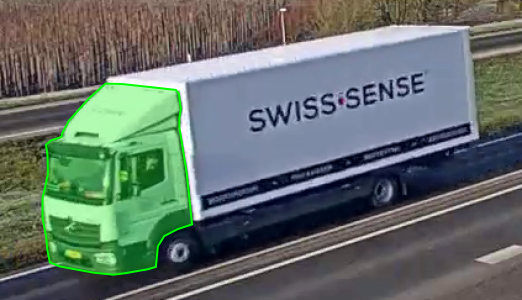} \\
                        Front Van &            931 &  \includegraphics[height=5mm]{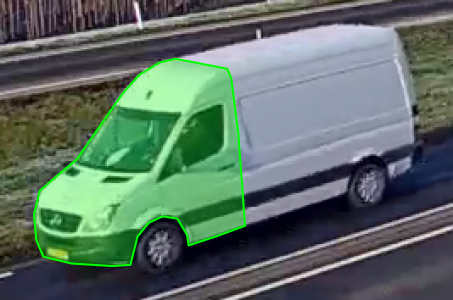} \\
                Load Camper Van &            255 &    \includegraphics[height=5mm]{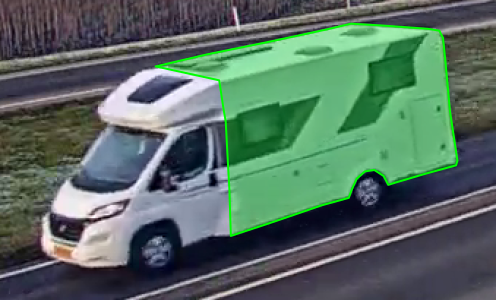} \\
                        Load Car &           1303 &   \includegraphics[height=5mm]{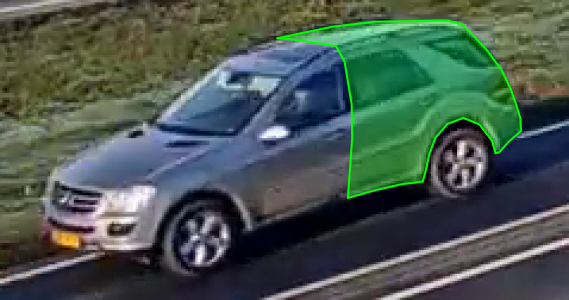} \\
                    Load Cuboid &            928 &    \includegraphics[height=5mm]{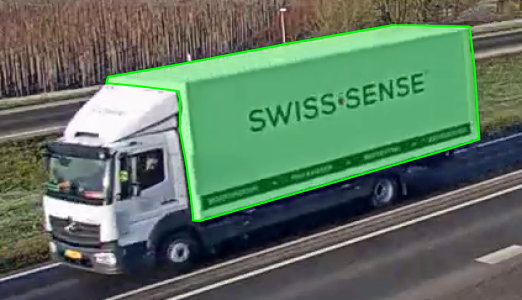} \\
                        Load Tank &            240 &  \includegraphics[height=5mm]{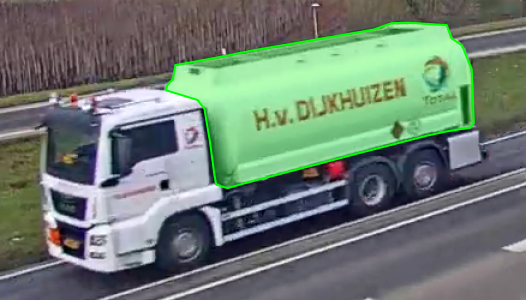} \\
                    Load Trough &            324 &    \includegraphics[height=5mm]{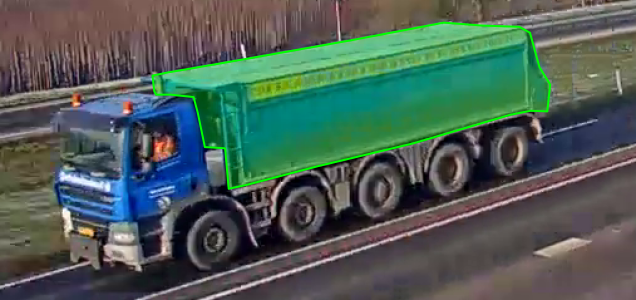} \\
                        Load Van &            235 &   \includegraphics[height=5mm]{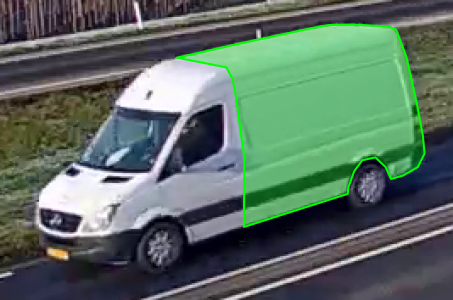} \\
                Roof Camper Van &            255 &    \includegraphics[height=5mm]{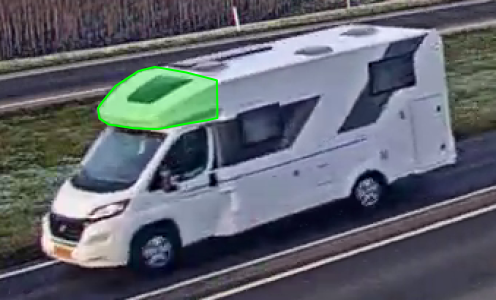} \\
    Roof Truck Car Transporter &             86 &     \includegraphics[height=5mm]{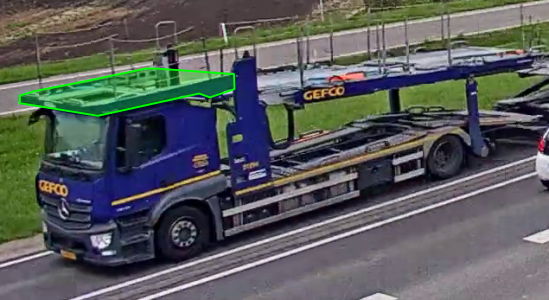} \\
                        Roof Van &            233 &   \includegraphics[height=5mm]{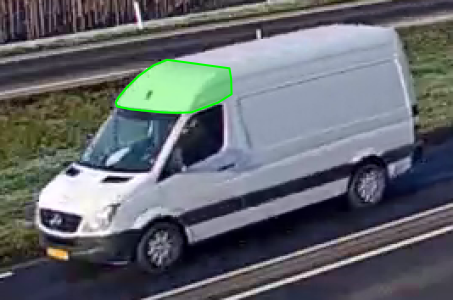} \\
    Support Truck Car Transporter &             86 &  \includegraphics[height=5mm]{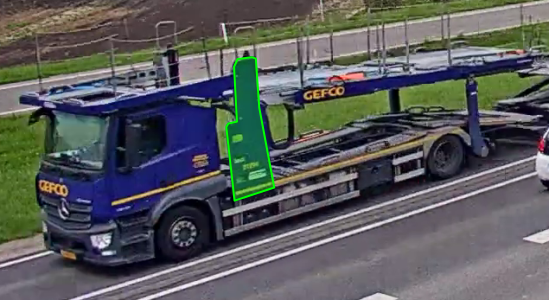} \\
                            Wheel &          12765 &  \includegraphics[height=5mm]{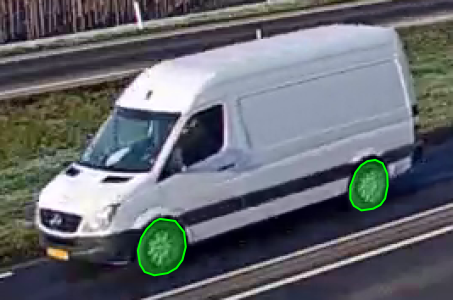} \\
\midrule
                    \textbf{Total} &  \textbf{21061} &                                  \\
\bottomrule
\end{tabular}
\end{table}

\subsection{Baseline CNN}\label{cnn}
For an independent assessment of the accuracy of our part-based classification approaches, an end-to-end CNN model was used as baseline, building on the architecture of our previous work that met the requirements of TLS8+1 A1 \cite{caduff2021exploring}. With this baseline we can monitor the progress and refinements of our current methods and provide an independent comparison with state-of-the-art black box CNN approaches.

The baseline model is built on a VGG-type architecture \cite{simonyan2014very}, specifically tailored and trained from scratch using the dataset represented in Table \ref{tab_categories}. The input resolution is 128 x 256 pixels, to achieve a good balance between computing power and the level of detail required for accurate classification. In fact, increasing the resolution further did not provide any significant improvement. 

The architecture comprises four convolutional layers, each followed by ReLU activation and max pooling, to efficiently extract and process the features from the input images. This is followed by two fully connected layers, also with ReLU activation, which perform the actual classification.

For the training, we used the Adam optimizer \cite{kingma2014adam} known for its effectiveness in handling sparse gradients and adaptive learning rate adjustments. The loss function is Cross Entropy, the standard choice for multi-class classification tasks. To improve the generalization capabilities of the model and prevent overfitting, we applied – as regularization techniques – batch normalization \cite{ioffe2015batch} for all, and in addition dropout \cite{srivastava2014dropout} but for the fully connected layers only.

A batch size of 32 was chosen, providing an optimal trade-off between memory usage and model update frequency. A learning rate of 0.001 (default starting values for Adam) was chosen to ensure steady convergence during training. 

\begin{figure}[]%
\centering
\includegraphics[width=72mm]{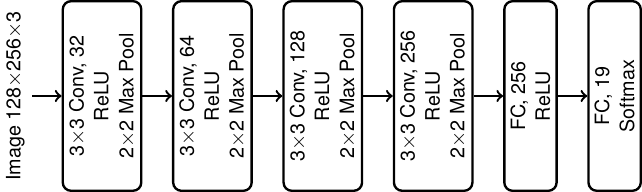}
\caption{CNN inference architecture used as baseline for the performance comparison with the new classification approaches.}\label{fig_cnn}
\end{figure}

Finally, early stopping was implemented as an additional regularization method that terminates the training process when the model’s performance on the validation set ceased to improve. The complete architecture and flow of the baseline CNN model are illustrated in Figure \ref{fig_cnn}.

\subsection{Classification Approach}\label{approach}

As shown conceptually in Figure \ref{fig_concept} our approach is composed of mainly three different steps, which will be explained in more detail below. Focus will be on the common points and the differences between the previous version (v1) \cite{caduff2022disentangling} using a hard decision tree and the present improvement (v2) based on spatial awareness in combination with a softmax regression approach. 

\begin{figure}[]
    \centering
    \includegraphics[width=53mm]{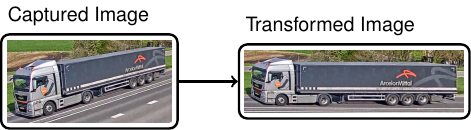}
    \caption{Camera calibration and transformation applied to a sample image to ensure consistency of the metric information over different camera views (Left: captured image. Right: calibrated and transformed).}\label{fig_trans}
\end{figure}

\textbf{Preprocessing (v1 and v2):} This represents an important step to ensure the consistency and reliability of our classification methods over different scenes i.e. camera views. During the feature construction step (Figure \ref{fig_concept}.) metric information is extracted, e.g. axle distance ratios for v1 or vehicle part positions for v2. This requires a) an identical scale along the x- and y-axes within a scene as well as b) a consistent global scale across different scenes. For a) a homography is defined aligning the side plane of the vehicle with the image sensor plane. The real-world coordinates required for the definition of this homography can be simply obtained from metric information on specific vehicle types (e.g. artic trucks as in Figure \ref{fig_trans}). This information also serves to define the global overall scale b). The effect of the transformation is illustrated in in Figure \ref{fig_trans}. It should be noted that a precision of ±10\% for the metric information is fully sufficient.

\begin{figure*}[]
    \centering
    \includegraphics[width=115mm]{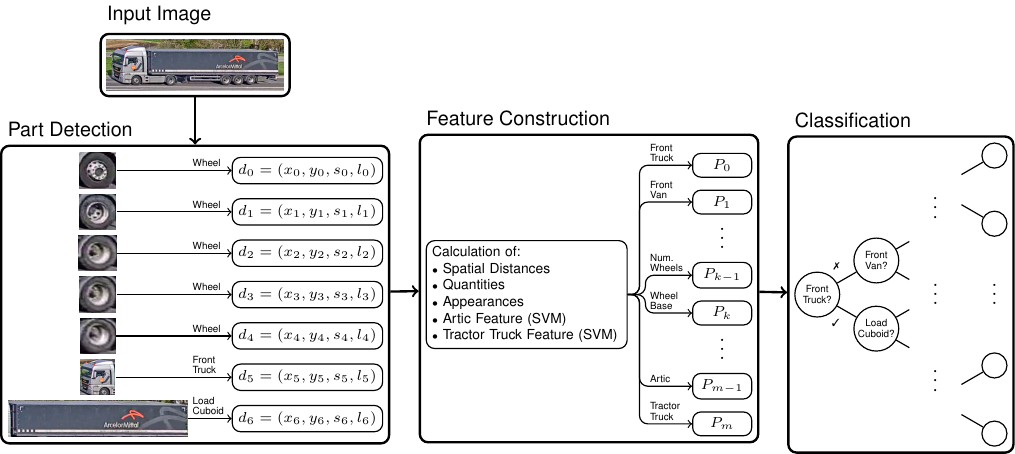}
    \caption{Workflow of the part-based decision tree classification approach: 1) using a CNN, semantically meaningful vehicle parts are detected with the information on location $\mathbf{x}^{(i)} = (x_i, y_i)^\text{T}$, confidence $s_i$, and category $l_i$, which are the basis for 2) the specific feature construction, followed by 3) an explainable decision tree classifier (for more detail see Section \ref{approach} and Figure \ref{fig_decision_tree_detail}).}\label{fig_decision_tree}
\end{figure*}

\textbf{Part Detection (v1 and v2):} For the detection of vehicle parts, a state-of-the-art DNN architecture of type YOLO was used \cite{redmon2016you}. It predicts both the bounding boxes and the categories of the vehicle parts in a single step. While not the most accurate in the spectrum of detection architectures \cite{zou2023object}, YOLO was selected for its real-time processing capabilities \cite{wang2022yolov7} – a crucial requirement for our ITS application. We use the Ultralytics implementation of YOLO5 \cite{ultralytics}, with an input resolution of 320 x 320 pixels (the S model). For the fine-tuning with the vehicle parts from table \ref{tab_parts} the hyperparameters were left at default including data augmentation settings. The part detection process, as conceptually shown in Figure \ref{fig_decision_tree}, outputs each detection $d_i$ with the part’s location $\mathbf{x}^{(i)} = (x_i, y_i)^{\text{T}}$, the confidence score $s_i$, and the predicted category $l_i$. These detections are then forwarded to the feature construction stage.

Following the preprocessing and part detection stages, common for both v1 and v2, we will now focus on the two different classification approaches v1 and v2. 

\textbf{Decision Tree Approach (v1):}
The full classification workflow of this method is presented in Figure \ref{fig_decision_tree}, with the two additional steps, feature construction and classification. 

The feature construction serves on the one hand to improve – based on specifically designed features – the robustness of the final classification step. In addition, using the same mechanism it allows for a simple extensibility of the model as was e.g. shown in \cite{caduff2022disentangling} on the basis of the so-called artic feature. The list of additionally constructed features and corresponding vehicle classes affected are given below:

\begin{figure*}[]
    \centering
    \includegraphics[width=97mm]{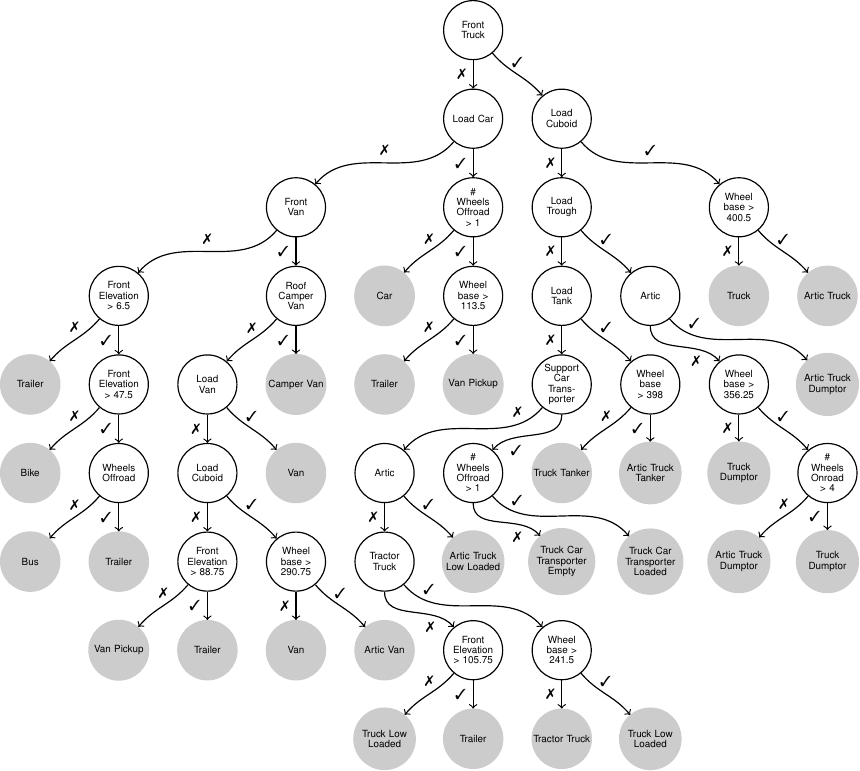}
    \caption{Visualization of the detailed decision tree classifier. All features needed for classification are shown in the nodes (white circles) and all categories are shown in the leafs (gray circles).}\label{fig_decision_tree_detail}
\end{figure*}

\begin{itemize}
    \item On-road and off-road wheels features: Determining categories by classifying wheels as on-road or off-road based on their position relative to the lowest wheel.
    \item Number of on-road and off-road wheels: Counting the wheels in each category to provide numerical data for classification.
    \item Vehicle wheelbase feature: Measuring the distance between the first and last on-road wheel to improve distinction between different vehicle lengths.
    \item Vehicle front elevation feature: Identifying the front of the driving vehicle or a loaded vehicle by measuring the distance from the first on-road wheel to the vehicle’s front.
    \item Artic feature (c.f. also \cite{caduff2022disentangling}): This feature is designed to distinguish between artic and non-artic subcategories, such as artic trucks versus regular trucks. It is constructed using three metric features: the scaled wheelbase, the distance ratio between the second and third on-road wheel, and the distance ratio between the third and fourth on-road wheel. These features make use of the characteristic wheel pattern of artic vehicles. The selection of the artic category is achieved with almost certainty (accuracy of 99.9\%) using a Radial Basis Function Kernel Support Vector Machine (RBF-SVM) with hyper-parameters C=0.1 and gamma=72. It is only applied to vehicles with more than three on-road wheels.
    \item Tractor truck feature: The tractor truck feature
    aims to distinguish between tractor trucks and trucks with low loading, which share the same vehicle parts. This feature is constructed from three metric features: scaled wheelbase, scaled front height, and the scaled number of on-road wheels. These metrics are utilized to define with almost certainty (accuracy of 99.4\%) the tractor truck category, using a RBF-SVM with hyperparameters C=5.2 and gamma=37.3 for classification. This feature is applied to vehicles with less than five on-road wheels. To ensure a balanced representation of tractor trucks in the SVM training this category is oversample to compensate for its relative rarity.
\end{itemize}

The final step of the classification pipeline is the decision tree, as illustrated in Figure \ref{fig_decision_tree_detail}. Input to the tree is the full set of final features consisting of the detected parts and the constructed features from step 2). These final features are represented in the nodes (white circles), leading down to the final vehicle categories in the leaves (gray circles). This structure allows for easy tracing of the classification path, offering clear insight into how decisions are made based on the input features.

As already noted above, despite its advantages, which are the explainability and simple extensibility, the v1 approach has a considerably drawback: a single false detection of a vehicle part will immediately lead to a wrong decision path due to its sequential decision-making in the decision tree. As an improvement, the present approach v2 with spatial awareness and softmax regresion was developed.

\begin{figure*}[]
\centering
\includegraphics[width=110mm]{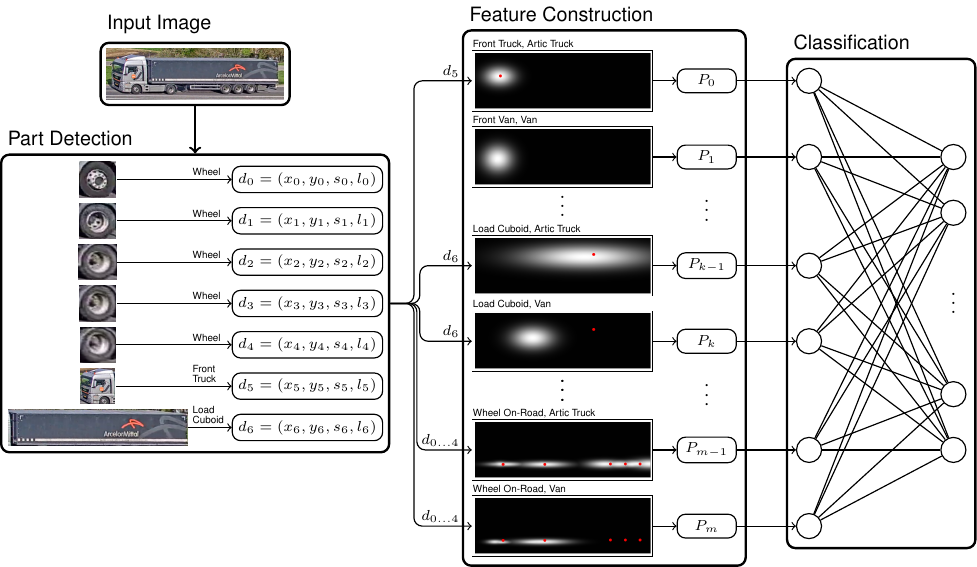}
\caption{Workflow of the part-based softmax regression classification approach: 1) using a CNN, semantically meaningful vehicle parts are detected with the information about location $\mathbf{x}^{(i)} = (x_i, y_i)^{\text{T}}$, confidence $s_i$, and category $l_i$, which are the basis for 2) the specific feature construction, followed by 3) a explainable linear regression classifier (for more detail see section \ref{approach} and figure \ref{fig_regression_detail}).}\label{fig_regression}
\end{figure*}

\begin{figure*}[]%
\centering
\includegraphics[width=138mm]{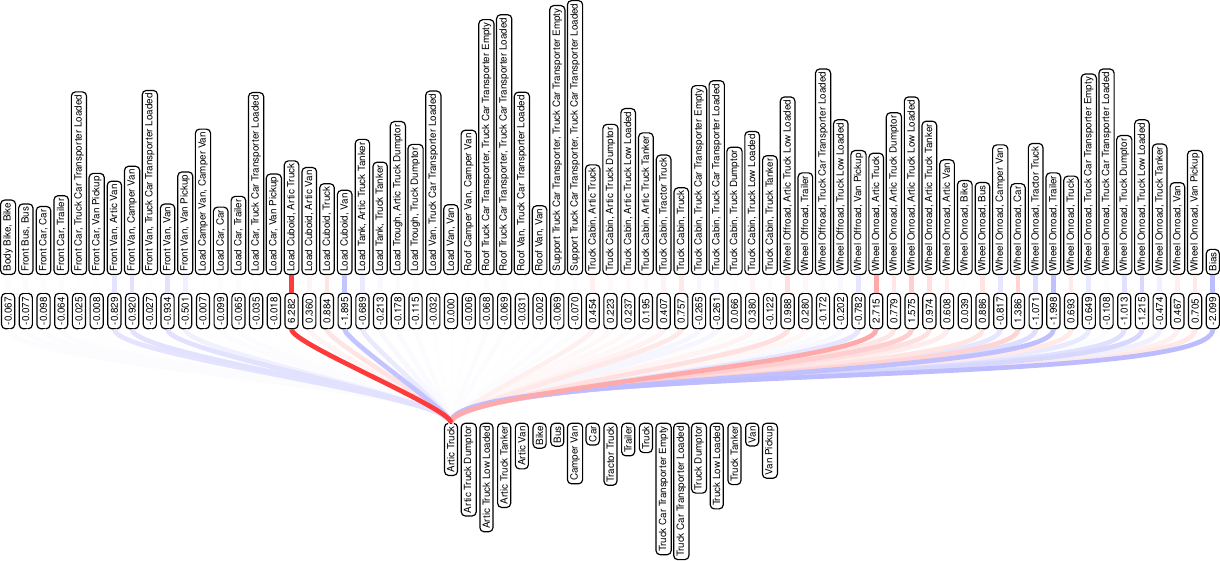}
\caption{Visualization of the softmax regression classification for the category Artic Truck. The top row displays all features utilized in the classification process. The middle row illustrates the weighted feature scores, with color indicating the weighting: red for positive and blue for negative contributions. The bottom row depicts the vehicle categories.}\label{fig_regression_detail}
\end{figure*}

\textbf{Softmax Regression Approach (v2):} The workflow of this approach is represented in Figure \ref{fig_regression} again with the two additional steps, feature construction and classification.

In the feature construction stage a so-called part score $P_k$ is calculated for all possible features $k$, with $k$ ranging from 1 to 69. Here, the list of features $k$ includes all combinations of vehicle parts with all vehicle categories in which the respective feature can occur (Figure \ref{fig_regression}, center and Figure \ref{fig_regression_detail}, top). The calculation of $P_k$ is schematically represented in Figure \ref{fig_regression} and formally consists of two steps as follows:

\begin{enumerate}
    \item Location score calculation (equation (\ref{eq_loc_score})):
    \begin{itemize}
        \item For every part detection $d_i$ at location $\mathbf{x}^{(i)}$, the location score $L_i^k$ is calculated for each feature $k$, for which part $i$ is relevant. Therefore all possible mean locations of this specific part $i$ for the vehicle in the given feature $k$ is taken into account (positions $\mathbf{x}_k^{(j)}$) and the maximum over all possible mean locations $\mathbf{x}_k^{(j)}$ is taken.
        \item The location score $L_i^k$ – representing a probability for the presence of part i at location $\mathbf{x}^{(i)}$ for the given vehicle category – is modelled using an unnormalized gaussian kernel function, with values ranging between 0 and 1. The gaussian kernel parameters, mean locations $\mathbf{x}_k^{(j)}$ and (diagonal) covariance matrices $\mathbf{\Sigma}_k^{(j)}$, are derived from fitting Gaussian Mixture Models (GMMs) to the training data. We found empirically that the multiplication of the covariance matrices $\mathbf{\Sigma}_k^{(j)}$ by a factor of four improved the convergence of the following softmax regression due to the fact, that detection further away from a mean locations  $\mathbf{x}_k^{(j)}$ still receive a non-zero location score. The number of modes for the GMM is determined using the Bayesian Information Criterion (BIC) for model selection.
        \item 	It should be noted that multiple instances of part detections (e.g. wheel detections $d_0$,…,$d_4$ in Figure \ref{fig_regression}) may contribute to the same feature $k$. Therefore, in a second step the final part score calculation is carried out.
    \end{itemize}
    \item Part score calculation (equation (\ref{eq_part_score})):
    \begin{itemize}
        \item 	The part score $P_k$ is calculated by multiplying the location score $L_i$ with the respective detection confidence score $s_i$ from the part detector and summing over all part instances contributing to this features $k$ ($n_k^{det}$). The sum is normalized by the number of expected detections for the respective part instance ($n_k^{exp}$).
    \end{itemize}
\end{enumerate}

\begin{equation} \label{eq_loc_score}
    L_i^k=\max_{j=1..n_k }e^{-\frac{1}{2} \left(\mathbf{x}^{(i)}-\mathbf{x}_k^{(j)}\right)^\text{T} \left(4\mathbf{\Sigma}_k^{(j)}\right)^{-1}\left(\mathbf{x}^{(i)}-\mathbf{x}_k^{(j)}\right)}
\end{equation}
\begin{equation} \label{eq_part_score}
    P_k = \frac{1}{n_k^{exp}}\sum_{j=1}^{n_k^{det}}s_i L_i^k
\end{equation}

The final result of the feature construction step consists of the list of part scores $P_k$ for each possible feature $k$ as e.g. represented in the top row of Figure \ref{fig_regression_detail}. 

In the classification stage, the constructed part score vector $P_k$ is fed into a softmax regression classifier, which linearly maps the features to the 19 fine-grained vehicle categories. The classifier is trained using the Adam optimizer \cite{kingma2014adam} and Cross Entropy Loss, with $\text{L}_2$ weight regularization set at 0.001. The batch size for training is set to 32, with a learning rate of 0.001. Early stopping is employed as an additional regularization technique.

This softmax regression approach brings several improvements over the decision tree approach. By processing all features in parallel and leveraging the full spatial relationship of vehicle parts, it anticipates more robust predictions, especially in case of false part detections. Unlike the heavily handcrafted features in the decision tree approach, the features in this method are derived systematically and processed in parallel, enhancing the overall prediction robustness.

Despite its sophistication, this approach maintains the simple extensibility and explainability crucial for ITS applications. The extensibility is now obtained by extension of the list of features k with the spatial probability maps of a new vehicle. This includes  country specific changes of certain vehicle classes as e.g. artic trucks, which may have different wheel patterns in the USA as compared to Europe. The explainability is obtained by visualizing and scoring the detected vehicle parts and the linear nature of the classification model. By examining the weighting of the features (part scores), one can interpret the classification results. For instance, Figure \ref{fig_regression_detail} demonstrates the weighted features for the detected parts of an artic truck, indicating which of the detected features strengthen (red) or weaken (blue) the classification.

\section{Results and Discussion}\label{results}

\begin{table*}[]
\caption{5-fold cross-validation mean ($\mu$) and standard error ($\sigma_{\mu}$) of the precision and recall values per vehicle part of the detector model (detection threshold 0.5). The column ”Objects” shows the average number of objects used for the evaluation.}\label{tab_resDetector}
\begin{center}
\begin{tabular*}{0.85\textwidth}{@{\extracolsep\fill}lrrrrr}
\toprule
& & \multicolumn{2}{@{}c@{}}{Precision} & \multicolumn{2}{@{}c@{}}{Recall}  \\\cmidrule{3-4}\cmidrule{5-6}%
                    Vehicle Part &  Objects &  $\mu$ &  $\sigma_{\mu}$ &  $\mu$ &  $\sigma_{\mu}$ \\
\midrule
                        Body Bike &            54.8 &           0.993 &                     0.006 &        0.993 &                  0.004 \\
                        Front Bus &            42.6 &           0.995 &                     0.004 &        0.972 &                  0.004 \\
                        Front Car &           260.4 &           0.989 &                     0.002 &        0.990 &                  0.002 \\
                        Front Van &           186.2 &           0.982 &                     0.002 &        0.992 &                  0.003 \\
                    Load Camper Van &            51.0 &           0.988 &                     0.006 &        0.988 &                  0.006 \\
                            Load Car &           260.6 &           0.988 &                     0.002 &        0.988 &                  0.002 \\
                        Load Cuboid &           185.6 &           0.986 &                     0.004 &        0.997 &                  0.002 \\
                        Load Tank &            48.0 &           0.975 &                     0.008 &        0.983 &                  0.006 \\
                        Load Trough &            64.8 &           0.985 &                     0.004 &        0.979 &                  0.003 \\
                            Load Van &            47.0 &           0.971 &                     0.009 &        0.975 &                  0.008 \\
                    Roof Camper Van &            51.0 &           0.981 &                     0.005 &        0.984 &                  0.006 \\
        Roof Truck Car Transporter &            17.2 &           0.977 &                     0.012 &        0.965 &                  0.012 \\
                            Roof Van &            46.6 &           0.970 &                     0.004 &        0.984 &                  0.008 \\
    Support Truck Car Transporter &            17.2 &           0.988 &                     0.010 &        0.988 &                  0.010 \\
                        Truck Cabin &           326.2 &           0.998 &                     0.001 &        0.998 &                  0.001 \\
                            Wheel &          2553.0 &           0.996 &                     0.000 &        0.997 &                  0.000 \\
\bottomrule
\end{tabular*}
\end{center}
\end{table*}

\begin{table*}[]
\caption{Mean ($\mu$) and standard error ($\sigma_{\mu}$) of the accuracy values of the 5-fold cross-validation overall (”All”) and for the individual vehicle categories. The column ”Samples” shows the average number of samples used for the evaluation.}\label{tab_resClassifier}
\begin{center}
\begin{tabular*}{\textwidth}{@{\extracolsep\fill}lrrrrrrr}
\toprule
& & \multicolumn{2}{@{}c@{}}{Softmax Reg.} & \multicolumn{2}{@{}c@{}}{Decision Tree} & \multicolumn{2}{@{}c@{}}{CNN}  \\\cmidrule{3-4}\cmidrule{5-6}\cmidrule{7-8}%
                Vehicle Category & Samples & $\mu$ & $\sigma_{\mu}$ & $\mu$ & $\sigma_{\mu}$ & $\mu$ &  $\sigma_{\mu}$ \\
\midrule
                            All &           955.8 &                   0.986 &                             0.001 &              0.985 &                        0.001 &    0.983 &              0.001 \\
                    Artic Truck &            91.6 &                   0.993 &                             0.006 &              1.000 &                        0.000 &    0.996 &              0.004 \\
            Artic Truck Dumptor &            17.2 &                   0.965 &                             0.013 &              0.976 &                        0.013 &    0.965 &              0.021 \\
            Artic Truck Low Loaded &             9.6 &                   0.920 &                             0.044 &              0.920 &                        0.044 &    0.918 &              0.034 \\
                Artic Truck Tanker &            10.0 &                   0.980 &                             0.018 &              0.980 &                        0.018 &    0.960 &              0.036 \\
                        Artic Van &             9.2 &                   0.978 &                             0.020 &              0.978 &                        0.020 &    0.956 &              0.024 \\
                            Bike &            54.4 &                   1.000 &                             0.000 &              1.000 &                        0.000 &    1.000 &              0.000 \\
                            Bus &            42.6 &                   0.972 &                             0.004 &              0.967 &                        0.005 &    0.981 &              0.008 \\
                            Car &           251.6 &                   0.994 &                             0.002 &              0.994 &                        0.002 &    0.993 &              0.002 \\
                        Trailer &            99.0 &                   0.998 &                             0.002 &              0.994 &                        0.002 &    0.990 &              0.006 \\
                            Truck &            62.6 &                   0.981 &                             0.005 &              0.984 &                        0.006 &    0.990 &              0.006 \\
    Truck Car Transporter Empty &             9.2 &                   1.000 &                             0.000 &              1.000 &                        0.000 &    0.978 &              0.020 \\
    Truck Car Transporter Loaded &             8.0 &                   0.975 &                             0.022 &              0.925 &                        0.045 &    0.975 &              0.022 \\
                    Truck Dumptor &            47.6 &                   0.975 &                             0.009 &              0.979 &                        0.008 &    0.987 &              0.004 \\
                Truck Low Loaded &            27.8 &                   0.935 &                             0.019 &              0.913 &                        0.022 &    0.899 &              0.026 \\
                    Truck Tanker &            38.0 &                   0.984 &                             0.009 &              0.979 &                        0.012 &    0.974 &              0.011 \\
                    Tractor Truck &             4.6 &                   0.790 &                             0.098 &              0.900 &                        0.089 &    0.830 &              0.106 \\
                            Van &            67.2 &                   0.988 &                             0.008 &              0.985 &                        0.010 &    0.985 &              0.006 \\
                        Van Pickup &            54.6 &                   0.985 &                             0.006 &              0.985 &                        0.006 &    0.974 &              0.008 \\
                        Camper Van &            51.0 &                   0.984 &                             0.007 &              0.980 &                        0.005 &    0.980 &              0.009 \\
\bottomrule
\end{tabular*}
\end{center}
\end{table*}

\begin{figure*}[]%
\centering
\includegraphics[width=150mm]{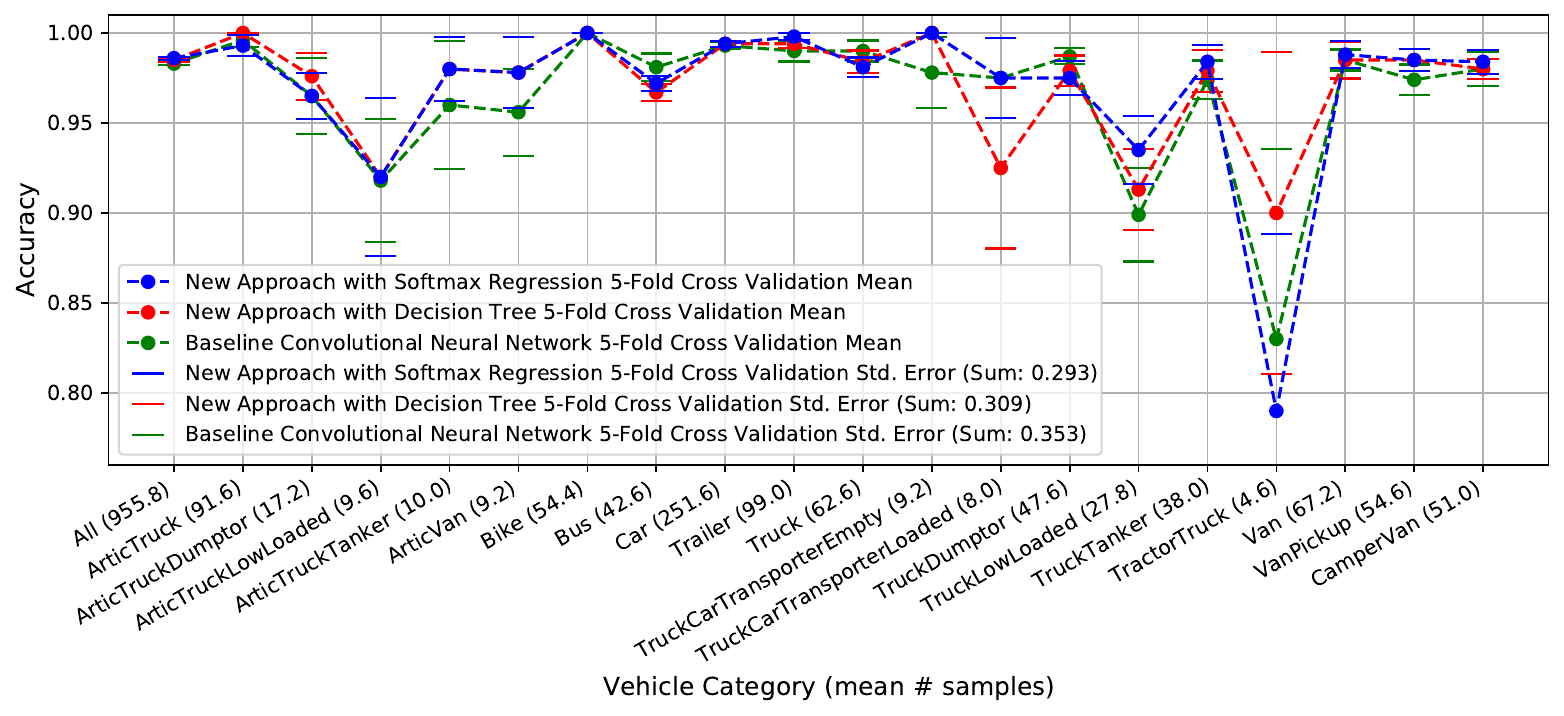}
\caption{Mean and standard error (summed over all categories parentheses) values of the accuracy of the 5-fold cross-validation overall (”All”) and for each of the 19 vehicle categories of the softmax regression approach. The decision tree approach and the baseline CNN results are shown for comparison. The detailed values can be found in Table \ref{tab_resClassifier}.}\label{fig_res_plot}
\end{figure*}

Below we compare the classification accuracies of the individual vehicle categories and the overall accuracy of the softmax regression approach with the decision tree approach and the baseline CNN. The results shown in Table \ref{tab_resClassifier} and Figure \ref{fig_res_plot}, represent the mean values and standard errors of the accuracy (individual categories and overall) from the 5-fold cross-validation process, compared across the three different approaches.

\textbf{Classification Accuracies:}
The softmax regression approach achieves an overall accuracy of 98.6\% across all categories, with individual category accuracies all above 92\%, except the single very rare “Tractor Truck” categories with an accuracy of 79\%. 

\textbf{Comparative Performance:}
To ensure a fair and consistent comparison, all components of all approaches were trained using the same 5-fold cross-validation data split. When the tree approaches are compared, the accuracies and variances across all vehicle categories reveal essentially identical performance profiles. Notably, the small differences in individual vehicle category accuracies systematically fall within the 5-fold cross-validation variability and thus below statistical significance. Furthermore, the accuracy curves of the different approaches qualitatively behave similarly across the various vehicle categories. This is obvious for the two part-based approaches (v1 and v2), due to identical basic part detector. However, for the baseline CNN this is far from being trivial and proves – a posteriori – that the decomposition of the CNN in a part detector 1) (in Figure \ref{fig_concept}) and subsequent classifiers 2) + 3) is as efficient as the baseline CNN. Here the backbone of the CNN is identified with 1) and the fully connected layers with 2) +3). These findings are in line with findings in related works \cite{zhou2014object}, which demonstrate that CNNs for scene classification automatically discover meaningful objects (here part) detectors . Notably, rare vehicle categories exhibit a drop in accuracy or higher variance in the 5-fold cross-validation, emphasizing the need for strategies to balance the training sample sizes of individual vehicle categories.

\textbf{Accuracy vs. Explainability:}
The near-identical accuracy between the part-based approaches and the CNN baseline might seem intuitive but is far from trivial. There is an ongoing discussion regarding the trade-off between accuracy and explainability in machine learning models \cite{minh2022explainable}. Some researchers argue that explainable models often sacrifice the accuracy that is inherent in black-box models, suggesting a possible inverse relationship between the two. Our results suggest that this inverse relationship is not necessarily observed.

\textbf{Part Detector Performance:}
The results of the part detector alone, presented in Table \ref{tab_resDetector}, show precision and recall values generally above 95\% for all parts. This high performance is crucial, as errors in the part detector directly impact the subsequent classification process, leading to potential misclassifications. Indeed, most misclassification cases in vehicle categories stem from inaccuracies in the part detection phase. Therefore, the robustness of the classification with respect to such part misclassification is essential.

\textbf{Robustness of Classification:}
The resilience of the softmax regression approach with respect to false vehicle part classification is highlighted through the results of Table \ref{tab_resRobust}. The performance of the softmax regression is compared to that of the decision tree approach under varying detection thresholds of the part detector. Lowering this threshold (“Thr.” in Table \ref{tab_resRobust}) from the default value 0.5 down to 0.001, i.e. making the detector extremely sensitive, emulates the case of false positive part detections. The decision tree approach demonstrates a pronounced degradation in classification accuracy. In comparison, the softmax regression approach demonstrates remarkable stability in classification accuracy across all thresholds, even when the detection threshold is significantly lowered. This robustness is illustrated in two scenarios: when the detection confidence score ($s_i$ in equation \ref{eq_part_score}) is retained (”Sc. retained”) and when it’s increased to a value of 0.5 (”Sc. adapted”) to emulate a high score for the corresponding detection. Only in the latter scenario, the softmax regression approach’s accuracy slightly decreases (by $\sim$ 0.7\%) with the lowest threshold but remains sufficiently high to fulfill ITS requirements. This marked difference in robustness is due to the fact that the softmax regression approach uses the spatial relationships among the detected vehicle parts and the parallel processing of the features. The advantage of the softmax regression approach in dealing with the inherent uncertainties and shortcomings of part detection in vehicle classification is of considerable importance for practical ITS applications where robustness is a key issue.

\textbf{Explainability of False Classifications:}
An advantage of the part-based approaches is their capacity to explain false classifications. For instance, a misclassification between an artic truck dumptor and an artic truck was analyzed and found to be caused by the misinterpretation of a dumptor’s tall trough as a cuboid load in the part detection phase. Such diagnostics are challenging with end-to-end CNN approaches, where understanding is typically post-hoc and based on tools like saliency maps.

\textbf{Processing Time:}
The only inconvenience of the part-based methods with respect to the baseline CNN approach is that the former require longer processing times due to the time taken by the part detector. For instance, on an embedded processing platform (Core i7- 6820EQ @ 2.8GHz), the part-based approach’s inference time is significantly slower than the CNN approach ( 25ms vs. 7ms), albeit still suitable for real-time processing – which is the main requirement for ITS applications.

\textbf{Future Improvements and Directions:}
The current version of the part-based classifier still leaves ample room for enhancement. Stages of the classification chain have not been systematically optimized and individual components were trained separately. Future work will look towards more holistic, end-to- end training, applying heuristics in the final classification step, such as considering mutually exclusive vehicle parts, and adopting more generic vehicle parts, such as more geometrically oriented vehicle fronts, to increase accuracy and reliability. Our ongoing development aims to refine the set of vehicle parts used and optimize the entire classification chain for high-precision classification suitable for ITS applications.

\begin{table}[]
\caption{Mean values of the accuracy of the 5-fold cross-validation overall for the decision tree and the softmax regression approach for different detection thresholds (”Thr.”). For the softmax regression approach, the ”Sc. retained” column shows the accuracy when the detection confidence score is retained and the ”Sc. adapted” column shows the accuracy when the detection confidence score is set to a minimum value of 0.5.}\label{tab_resRobust}
\begin{tabular}{rrrr}
\toprule
 & & \multicolumn{2}{@{}c@{}}{Softmax Reg.}  \\\cmidrule{3-4}
    Thr. &  Decision Tree &  Sc. retained &  Sc. adapted \\
\midrule
        0.500 &          0.985 &               0.986 &                          0.986 \\
        0.100 &          0.981 &               0.986 &                          0.986 \\
        0.010 &          0.971 &               0.986 &                          0.983 \\
        0.001 &          0.929 &               0.986 &                          0.980 \\
\bottomrule
\end{tabular}
\end{table}

\section{Conclusion}\label{conclusion}
The search for robust, accurate, flexible and interpretable classification systems in ITS has led us to develop new part-based approaches as an alternative to traditional end-to-end CNNs. These methods disentangle the CNN into a more manageable framework, focusing on detection of concrete and interpretable vehicle parts refined through feature construction and finalized with a simple, explainable classifier. Our research has finally resulted in a framework for fine-grained vehicle category classification that not only enhances interpretability but also ensures easy extensibility, allowing for the addition of new vehicle categories without retraining the intensive part detector.

The part detector employs a state-of-the-art DNN, specifically YOLOv5, trained on 16 semantically strong vehicle parts. The annotated training dataset was prepared using an efficient, semi-automated iterative approach, including a manual correction step, and bootstrapped with a few-shot model trained on vehicle parts. Following the part detector, two distinct classification approaches emerged: the former decision tree approach and the present softmax regression approach. The decision tree approach, while effective, showed pronounced sensitivity to false detections, which lead us to the development of the softmax regression approach. This approach includes the information on the spatial relationship of the vehicle parts and processes the resulting features in parallel, significantly enhancing robustness.

Our part-based methods have demonstrated comparable accuracy to state-of-the-art end-to-end CNNs, challenging the presumed trade-off between accuracy and explainability in the research field of XAI.

While promising, our approaches come with challenges and limitations. The performance depends heavily on the precision of part detection and may vary under different environmental or operational conditions. Thus, potential for further development lies in refining the part detector, enhancing the identification of spatial relationships, and optimizing the classification chain. Our goal is to create a classifier that not only understands a wide array of vehicle categories but does so with an intuitive sense of the underlying semantics. 

Our research may have significant implications for ITS applications. Improved vehicle classification directly impacts traffic monitoring, safety measures, and efficiency, contributing to smarter, more responsive transportation networks. The adaptability and accuracy of our methods promise to support the evolving needs of urban development and traffic management, marking a step towards more integrated and intelligent systems.

\vspace{0.5cm}
\noindent\textbf{Acknowledgments:}
This project was supported by the Swiss Innovation Agency Innosuisse.


\bibliographystyle{ieeetr}%
\bibliography{sn-bibliography}

\end{document}